\documentclass[sigconf,natbib=false]{acmart}

\usepackage[all]{xy}
\usepackage{algorithm}
\usepackage{flushend}
\usepackage{paralist}
\usepackage{algpseudocode}
\makeatletter
\algnewcommand{\LineComment}[1]{\Statex \hskip\ALG@thistlm \(\triangleright\) #1}
\gdef\@copyrightpermission{
  \begin{minipage}{0.3\columnwidth}
   \href{https://creativecommons.org/licenses/by/4.0/}{\includegraphics[width=0.90\textwidth]{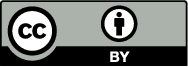}}
  \end{minipage}\hfill
  \begin{minipage}{0.7\columnwidth}
   \href{https://creativecommons.org/licenses/by/4.0/}{This work is licensed under a Creative Commons Attribution International 4.0 License.}
  \end{minipage}
  \vspace{5pt}
}
\makeatother
\usepackage{graphicx}
\graphicspath{{images/}}

\AtBeginDocument{%
  }

\copyrightyear{2024}
\acmYear{2024}
\setcopyright{rightsretained}
\acmConference[AISec '24]{Proceedings of the 2024 Workshop on Artificial Intelligence and Security }{October 14--18, 2024}{Salt Lake City, UT, USA}
\acmBooktitle{Proceedings of the 2024 Workshop on Artificial Intelligence and Security (AISec '24), October 14--18, 2024, Salt Lake City, UT, USA}
\acmDOI{10.1145/3689932.3694767}
\acmISBN{979-8-4007-1228-9/24/10}

\RequirePackage[
  datamodel=acmdatamodel,
  style=acmnumeric,
  ]{biblatex}

\begin{document}

\title{It's Our Loss: No Privacy Amplification for Hidden State DP-SGD With Non-Convex Loss}

\author{Meenatchi Sundaram Muthu Selva Annamalai}
\email{meenatchi.annamalai.22@ucl.ac.uk}
\affiliation{%
  \institution{University College London}
  \city{London}
  \country{United Kingdom}
}

\renewcommand{\shortauthors}{Meenatchi Sundaram Muthu Selva Annamalai}

\begin{abstract}
  Differentially Private Stochastic Gradient Descent (DP-SGD) is a popular iterative algorithm used to train machine learning models while formally guaranteeing the privacy of users.
  However, the privacy analysis of DP-SGD makes the unrealistic assumption that all intermediate iterates (aka internal state) of the algorithm are released since, in practice, only the final trained model, i.e., the final iterate of the algorithm is released.
  In this hidden state setting, prior work has provided tighter analyses, albeit only when the loss function is constrained, e.g., strongly convex and smooth or linear.
  On the other hand, the privacy leakage observed empirically from hidden state DP-SGD, even when using non-convex loss functions, suggests that there is in fact a gap between the theoretical privacy analysis and the privacy guarantees achieved in practice. 
  Therefore, it remains an open question whether hidden state privacy amplification for DP-SGD is possible for all (possibly non-convex) loss functions in general.

  In this work, we design a counter-example and show, both theoretically and empirically, that a hidden state privacy amplification result for DP-SGD for all loss functions in general is not possible.
  By carefully constructing a loss function for DP-SGD, we show that for specific loss functions, the final iterate of DP-SGD alone leaks as much information as the sequence of all iterates combined.
  Furthermore, we empirically verify this result by evaluating the privacy leakage from the final iterate of DP-SGD with our loss function and show that this exactly matches the theoretical upper bound guaranteed by DP.
  Therefore, we show that the current privacy analysis for DP-SGD is tight for general loss functions and conclude that no privacy amplification is possible for DP-SGD in general for all (possibly non-convex) loss functions.
\end{abstract}

\ccsdesc[500]{Security and privacy~Privacy-preserving protocols}

\keywords{Differential Privacy; Machine Learning; DP-SGD}

\maketitle

\section{Introduction}
Machine learning models trained using the stochastic gradient descent (SGD) algorithm have been known to leak potentially sensitive information about the training dataset~\cite{shokri2017membership,carlini2022membership,hayes2017logan}.
To prevent this, a modified version of SGD, called Differentially Private Stochastic Gradient Descent (DP-SGD)~\cite{abadi2016deep} is used to train models privately.
DP-SGD clips the gradients of each individual data point and adds carefully calibrated noise so that the DP-SGD algorithm satisfies formal Differential Privacy (DP)~\cite{dwork2006calibrating} guarantees.
Informally, DP bounds the information leakage from an algorithm up to a privacy parameter $\varepsilon$, thus preventing any adversary from accurately learning sensitive information about the training dataset.
Previously, DP-SGD required prohibitively large noise scales in order to enjoy reasonable levels of privacy guarantees.
However, tighter privacy analyses~\cite{kairouz2015composition,mironov2017renyi} and privacy amplification results~\cite{bassily2014private,abadi2016deep,balle2018privacy} have significantly reduced the magnitude of noise necessary, thus making DP-SGD much more practical in recent years.

One such amplification result that is an active area of research is \textit{hidden state privacy amplification}.
Put simply, DP-SGD is an iterative algorithm that updates some initial model parameters $\theta_0$ over $T$ steps, outputting only the final iterate $\theta_T$.
However, theoretically analyzing the privacy leakage from just the final iterate has been difficult in prior work~\cite{nasr2021adversary}.
Therefore, the state-of-the-art privacy analysis of DP-SGD assumes that the intermediate iterates $\theta_1,...,\theta_T$ are released as well.
This raises the question of whether the privacy analysis of DP-SGD can be improved further when this aspect is taken into account, i.e., whether the privacy guarantees of DP-SGD can be amplified given that the state (intermediate iterates) are hidden.

Indeed, prior work has provided tighter guarantees for DP-SGD in the hidden state setting, albeit only for constrained loss functions, e.g., strongly convex and smooth loss~\cite{ye2022differentially,chourasia2021differential} or linear loss~\cite{choquette-choo2024privacy}.
However, commonly used non-convex loss functions in modern deep learning models do not satisfy the necessary constraints for hidden state privacy amplification.
Therefore, the constraints on loss functions have been a significant limitation of prior work, which has left most differentially private machine learning methods unaffected by the existence of such hidden state privacy amplification results.

On the other hand, empirical results~\cite{nasr2023tight,cebere2024tighter,andrew2023one,nasr2021adversary,cherubin2024closed} have long observed that the privacy guarantees achieved by the final iterate of DP-SGD even with commonly used non-convex loss functions are much higher than those guaranteed by DP-SGD's theoretical privacy analysis.
This has led prior work to conjecture that the privacy analysis of DP-SGD can in fact be substantially improved when only the final iterate of DP-SGD is released, even for non-convex loss functions in general~\cite{nasr2023tight,cebere2024tighter}.
However, without any proof (or counter-example) of such a result, it remains an open research question whether hidden state privacy amplification for DP-SGD is possible for all loss functions in general.

\paragraph{Contributions.}
In this work, we design a counter-example and show, both theoretically and empirically, that a hidden state privacy amplification result for DP-SGD for all loss functions in general is not possible.
To that end, we first carefully construct a worst-case non-convex loss function for DP-SGD where the information of all previous iterates is encoded into the final iterate.
Theoretically, this means that the privacy leakage from the final iterate of DP-SGD with our worst-case loss function is exactly the same as the privacy leakage from all of the intermediate iterates of DP-SGD combined. 
Additionally, we verify this result empirically through auditing and find that the empirical privacy leakage from the final iterate of DP-SGD with our loss function matches the theoretical upper bound guaranteed by DP-SGD's current state-of-the-art privacy analysis exactly.

Our results show that without any constraints on the loss function, DP-SGD's current privacy analysis is indeed tight, even when only the final iterate is released.
Furthermore, our results are \textit{constructive} as we provide a concrete non-convex loss function that results in the same level of privacy leakage for the final iterate as for the sequence of all iterates.
Therefore, we confidently conclude that the privacy guarantees of DP-SGD cannot be improved further in the hidden state setting for all loss functions in general.

\section{Related Work}
\subsection{Hidden state privacy amplification}
Hidden state privacy amplification is a relatively new area of research.
Feldman et al.~\cite{feldman2018privacy} first introduced this idea under the moniker ``privacy amplification by iteration''.
Specifically, they showed that the privacy analysis of learning a model privately over one single training epoch could be tightened when only the last iterate of the epoch is released and the loss function is smooth and convex.
Subsequently, Choursaia et al.~\cite{chourasia2021differential} and Ye et al.~\cite{ye2022differentially} extended the amplification bound to training over multiple epochs when the loss function is strongly convex and smooth.
Separately, Choquette-Choo et al.~\cite{choquette-choo2024privacy} stated that for linear loss functions, the privacy guarantees provided by the final iterate of DP-SGD are equivalent to the Gaussian mechanism with random sensitivity $Binom(T, q)$ and variance $T\sigma^2$ and tightly analyze this mechanism using the Privacy Loss Distribution approach.

Thus far, the hidden state privacy amplification results have constrained the loss function in various ways, but the non-convex loss functions used in practice do not generally satisfy these constraints.
Therefore, it has been unclear whether these hidden state privacy amplification results can be extended to all loss functions in general.
However, in our work, we show that this is impossible by constructing a worst-case loss function where the privacy leakage from the final iterate of DP-SGD matches the theoretical privacy leakage of DP-SGD when all intermediate iterates are released.

\subsection{Auditing DP-SGD}
Hidden state DP-SGD is often referred to as DP-SGD under the ``black-box'' threat model, as in both cases, only the final iterate of DP-SGD is released.
Under this threat model, Jayaraman and Evans~\cite{jayaraman2019evaluating} audited DP-SGD and found that there is a large gap between the empirical privacy leakage observed and the theoretical upper bound guaranteed by DP.
Jagielski et al.~\cite{jagielski2020auditing} closed this gap slightly by using data poisoning and constant initial model parameters $\theta_0$ instead of randomly initializing them.
Yet, the empirical privacy leakage observed was still far from the theoretical upper bounds guaranteed.

Nasr et al.~\cite{nasr2021adversary} used a stronger, ``white-box'' threat model instead to audit DP-SGD and were the first to achieve empirical privacy leakages that matched the theoretical upper bounds, albeit only for worst-case neighboring datasets.
Essentially, the threat model considered by Nasr et al. is equivalent to releasing all intermediate iterates of DP-SGD.
Nasr et al. also considered the hidden state (``black-box'') setting but failed to achieve tight empirical estimates.
For \textit{natural} (average-case) neighboring datasets, Nasr et al.~\cite{nasr2023tight} achieved tight empirical privacy leakage estimates, but again only in the ``white-box'' threat model.
Therefore, they concluded that there is a gap between the theoretical guarantees provided by DP and the empirical privacy leakage that can be achieved when only the final iterate is released.

In recent work, De et al.~\cite{de2022unlocking}, Galen et al.~\cite{andrew2023one}, and Cebere et al.~\cite{cebere2024tighter} all audited the final iterate of DP-SGD under various settings (centralized and federated machine learning) and found that the empirical privacy leakage observed always falls short of the theoretical upper bounds guaranteed by DP.
Interestingly, when there is no sub-sampling, both Cebere et al.~\cite{cebere2024tighter}, and separately, Annamalai et al.~\cite{annamalai2024nearly} showed that the empirical privacy leakage observed for the final iterate of DP-SGD closely matches the theoretical guarantees.
Lastly, Cherubin et al.~\cite{cherubin2024closed} evaluated the empirical privacy leakage of DP-SGD using a new approach referred to as the ``Bayes Security measure''.
However, they fell short of applying their approach to the setting where only the final iterate is released as well.

These results, all put together, seem to suggest that the privacy analysis of DP-SGD can be improved when considering the setting where only the final iterate is released.
However, prior work auditing DP-SGD only considered loss functions commonly used in differentially private machine learning, even though the privacy guarantees of DP-SGD hold for worst-case loss functions as well.
In contrast, we construct a worst-case loss function in our work and show that the privacy analysis of DP-SGD is tight and cannot be improved for this loss function even when only the final iterate is released.

\section{Background}
In this section, we introduce the concepts of differential privacy, DP-SGD, trade-off functions, and auditing.
\subsection{Differential Privacy (DP)}
\begin{definition}[Differential Privacy (DP)~\cite{dwork2006calibrating}]
  \label{def:dp}
  A randomized mechanism $\mathcal{M} : \mathcal{D} \rightarrow \mathcal{R}$ is $(\varepsilon, \delta)$-differentially private if for any two neighboring datasets $D, D' \in \mathcal{D}$ and $S \subseteq \mathcal{R}$, it holds:
  \begin{equation*}
    \Pr[\mathcal{M}(D) \in S]  \leq e^\varepsilon \Pr[\mathcal{M}(D') \in S] + \delta
  \end{equation*}
\end{definition}

Informally, DP guarantees an information-theoretic upper bound (up to the privacy parameter $\varepsilon$) on any adversary's ability to distinguish between the output of $\mathcal{M}$ run on two neighboring inputs --- i.e., two datasets ($D, D'$) with a single record inserted/deleted.

\begin{theorem}[Advanced Composition~\cite{kairouz2015composition}]
  \label{thm:comp}
  Let $\mathcal{M}$ be a sequence of $(\varepsilon, \delta)$-DP mechanisms, i.e., $\mathcal{M} = (\mathcal{M}_1, \mathcal{M}_2, ..., \mathcal{M}_k)$, where each $\mathcal{M}_i$ can be chosen adaptively.
  Then for all $\delta' \geq 0$, $\mathcal{M}$ satisfies $(\tilde{\varepsilon}, \tilde{\delta})$-DP for $\tilde{\varepsilon} = \varepsilon \sqrt{2k \log(1 / \delta')} + k\varepsilon\frac{e^\varepsilon - 1}{e^\varepsilon + 1}$ and $\tilde{\delta} = k\delta + \delta'$.
\end{theorem}

The \textit{advanced composition theorem} shown above is an important theorem satisfied by DP that allows the outputs of multiple DP mechanisms to be combined without completely breaking the guarantees provided by DP.

\subsection{DP-SGD}
Differentially Private Stochastic Gradient Descent (DP-SGD)~\cite{abadi2016deep} is a popular algorithm used to train machine learning models with DP guarantees.
DP-SGD takes as input (1) the dataset $D$, (2) loss function $\ell$, (3) initial model parameters $\theta_0$, (4) learning rate $\eta$, (5) gradient clipping norm $C$, (6) noise multiplier $\sigma$, (7) sampling rate $q$, and (8) number of steps $T$ and outputs $\theta_T$ after applying the following update rule iteratively:
\begin{equation*}
  \theta_{k + 1} \leftarrow \theta_{k} - \eta \left( \sum_{x \in S_q(D)} \text{clip}_C(\nabla \ell(x; \theta_{k})) + \mathcal{N}(0, C^2 \sigma^2)\right)
\end{equation*}

Typically, $S_q$ is the Poisson sub-sampling operator, $C$ is set to 1, and $\sigma$ is calibrated appropriately such that DP-SGD satisfies $(\varepsilon, \delta)$-DP.
Observe that the DP guarantees hold for \textit{for any loss function} $\ell$ since the clip function enforces the sensitivity regardless of the loss function.
In this work, we abstract away the details of DP-SGD and write it as $\text{DP-SGD}(D; \ell, \theta_0, \eta, C, \sigma, q, T)$.
When there is no ambiguity in the hyper-parameters, we write it as $\text{DP-SGD}(D; \cdot)$.

\paragraph{Hidden State Privacy Amplification}
Although DP-SGD only outputs the final model $\theta_T$ (hidden state), in general, the privacy analysis of DP-SGD depends on the composition theorem (Theorem~\ref{thm:comp}), which assumes that \textit{all intermediate model parameters} $\theta_1, ..., \theta_T$ are released by the mechanism.
In previous work~\cite{ye2022differentially,chourasia2021differential,choquette-choo2024privacy}, the privacy analysis of DP-SGD in the hidden state setting has been tightened, but only when the loss function is constrained.
The latest of these results is presented by Choquette-Choo et al.~\cite{choquette-choo2024privacy}, who state that when the loss function is linear, the privacy guarantees of hidden state DP-SGD (with noise multiplier $\sigma$, sampling rate $q$, and $T$ steps) is equivalent to that of a Gaussian mechanism with random sensitivity $Binom(T, q)$ and variance $T\sigma^2$.
However, for general loss functions, no such privacy amplification has been proven, although such amplification is thought to be possible based on empirical results~\cite{nasr2023tight,cebere2024tighter,andrew2023one,nasr2021adversary,cherubin2024closed}.

\subsection{Trade-off functions}
\label{sec:tradeoff}
Implicit in the definition of DP is an information-theoretic limit on the adversary's ability to distinguish between outputs of a mechanism on neighboring inputs.
This limit can be expressed through the following hypothesis testing problem: Given some output $\theta$ of a DP mechanism $\mathcal{M}$ on neighboring inputs $D$ or $D'$
\begin{align*}
  H_0 &: \theta\text{ is drawn from }\mathcal{M}(D) \\
  H_1 &: \theta\text{ is drawn from }\mathcal{M}(D')
\end{align*}

Any adversary attempting to distinguish between $H_0$ and $H_1$ will achieve a False Positive Rate (FPR) and False Negative Rate (FNR).
DP guarantees that the achievable FPRs ($\alpha$) and FNRs ($\beta$) are bounded, which is characterized by a \textit{trade-off function}.

\begin{definition}[Trade-off function~\cite{dong2019gaussian}]
  \label{def:tradeoff}
  For any two probability distributions $P$, $Q$ on the same space, the trade-off function $T(P, Q): [0, 1] \rightarrow [0, 1]$ is defined as follows:
  \begin{equation*}
    T(P, Q)(\alpha) = \inf_{\phi} \{\beta_\phi : \alpha_\phi \leq \alpha \}
  \end{equation*}
  where the infimum is taken over all possible rejection rules $\phi$.
\end{definition}

Note that the most optimal test that achieves the smallest FNR is given by the Neyman-Pearson lemma~\cite{neyman1933ix}, which corresponds to the likelihood ratio test.

\begin{definition}[Likelihood Ratio Test~\cite{neyman1933ix}]
  \label{def:lira}
  For a given hypothesis test with null hypothesis $H_0: \theta \sim P$ and alternate hypothesis $H_1: \theta \sim Q$, the optimal test achieving the lowest FNR at a fixed FPR is given by thresholding the output of the following function:
  \begin{equation*}
    \Lambda(x) = \frac{p(x | Q)}{p(x | P)}
  \end{equation*}
  where $p(x | P)$ and $p(x | Q)$ are the probability density functions of $P$ and $Q$, respectively.
\end{definition}

\paragraph{Approximating trade-off function}
While the trade-off function for some simple mechanisms like the Laplace Mechanism and Gaussian Mechanism have closed-form expressions~\cite{dong2019gaussian}, the trade-off function for more complex mechanisms like DP-SGD (with sub-sampling and composition) has to be approximated.
To do so, we follow Nasr et al.'s approach~\cite{nasr2023tight} and use the ``Privacy Loss Distribution (PLD)''~\cite{koskela2020computing} of DP-SGD.
In this work, we abstract away the details of the approximation and simply write $\beta \leftarrow \text{PLD}(\varepsilon)(\alpha)$ to indicate the FNR predicted by the trade-off approximation at a given FPR using the PLD for DP-SGD (with composition) at a theoretical privacy level of $\varepsilon$.
Note that the approximated trade-off function will be \textit{symmetric in the neighboring datasets}, i.e., it will characterize the lowest FNR achievable regardless of whether the null hypothesis ($H_0$) is ``$\theta\text{ is drawn from }\mathcal{M}(D)$'' or ``$\theta\text{ is drawn from }\mathcal{M}(D')$''.

\subsection{Auditing DP}
Auditing is the process of \textit{empirically} verifying that the \textit{theoretical} guarantees provided by DP hold in practice.
Two main reasons that this might not happen are: (1) the privacy analysis of the mechanism can be improved further~\cite{nasr2021adversary} or (2) there are bugs in the implementation of the mechanism~\cite{tramer2022debugging,nasr2023tight}.
In this work, we are interested in investigating the former.
Regardless, the process of auditing remains the same.

Firstly, the mechanism $\mathcal{M}$ is run repeatedly on neighboring datasets $D$, $D'$ at a given level of privacy $\varepsilon$.
Next, the adversary tries to distinguish between the outputs of $\mathcal{M}(D)$ and $\mathcal{M}(D')$, resulting in an FPR and FNR.
Although confidence intervals for FPR and FNR are typically computed so that bugs can be identified with an associated level of confidence, in this work, we forgo this step to achieve the tightest possible guarantees.
Lastly, the FPR and FNR are converted into an empirical estimate for the level of privacy $\varepsilon_{emp}$ using the trade-off function of $\mathcal{M}$ (see Section~\ref{sec:audit}).

If the empirical estimate matches the expected theoretical guarantees, i.e., $\varepsilon_{emp} \approx \varepsilon$, the empirical privacy leakage we observe matches the theoretical upper bound guaranteed by DP.
Therefore, we can conclude that the privacy analysis of $\mathcal{M}$ is tight and cannot be improved further.
Otherwise, if the empirical estimate falls short of the expected theoretical guarantee, i.e., $\varepsilon_{emp} \ll \varepsilon$, the empirical privacy leakage observed is much lower than the theoretical upper bound.
This indicates that either (a) the adversary can be improved to better distinguish between the outputs, or (b) the theoretical privacy analysis can be improved further (e.g., via possible privacy amplification theorems).

\section{Our Loss Function}
We begin by providing an overview of how we construct our loss function.
First, we derive the likelihood ratio test, which is the optimal test to distinguish between $\text{DP-SGD}(D; \cdot)$ and $\text{DP-SGD}(D'; \cdot)$ when all model iterates are released.
Next, we construct a \textit{non-convex} loss function that performs this test at each iterate and encodes the result into the next iterate.
Then, we show that distinguishing between the final iterate is equivalent to distinguishing between the sequence of iterates when using our loss function.
That is, we show that an adversary with access to \textit{just the final iterate} of DP-SGD can distinguish between $D$ and $D'$ just as easily as an adversary with access to all intermediate iterates combined.
Crucially, \textit{the loss function is the only part of DP-SGD that we define} and we do not modify any other part of DP-SGD (unlike ``white-box'' auditing techniques~\cite{nasr2021adversary,nasr2023tight}).
Lastly, we explain how we evaluate the empirical privacy leakage from the final iterate of DP-SGD and compare it with the theoretical privacy guarantee through auditing.

For simplicity, we shall assume that $\eta = C = 1$ and that datasets are one-dimensional, i.e., $D \in \mathbb{R}^n$, but note that our construction is generic and can be modified accordingly.

\subsection{The likelihood ratio test}
Here, we introduce the likelihood ratio test when DP-SGD releases all iterates.
In this setting, distinguishing between $\text{DP-SGD}(D; \cdot)$ and $\text{DP-SGD}(D'; \cdot)$ reduces to distinguishing between $\prod_{i = 1}^T \mathcal{N}(0, \sigma^2)$ and $\prod_{i = 1}^T \begin{cases} \mathcal{N}(1, \sigma^2)\text{ w.p. }q \\ \mathcal{N}(0, \sigma^2)\text{ w.p. }1 - q\end{cases}$.
We know that the optimal test is derived by thresholding the output of the following likelihood ratio function from the Neyman-Pearson lemma~\cite{neyman1933ix} where $\theta = (\theta_1,...,\theta_T)$:
\begin{align*}
  \Lambda(\theta) &= \frac{\Pr\left[\theta | \prod_{i = 1}^T \begin{cases} \mathcal{N}(1, \sigma^2)\text{ w.p. }q \\ \mathcal{N}(0, \sigma^2)\text{ w.p. }1 - q\end{cases}\right]}{\Pr[\theta | \prod_{i = 1}^T \mathcal{N}(0, \sigma^2)]} \\
  &= \frac{\prod_{i = 1}^T \Pr\left[\theta_i | \begin{cases} \mathcal{N}(1, \sigma^2)\text{ w.p. }q \\ \mathcal{N}(0, \sigma^2)\text{ w.p. }1 - q\end{cases}\right]}{\prod_{i = 1}^T \Pr[\theta_i | \mathcal{N}(0, \sigma^2)]} \\
  &= \prod_{i = 1}^T \frac{q\Pr[\theta_i | \mathcal{N}(1, \sigma^2)] + (1 - q)\Pr[\theta_i | \mathcal{N}(0, \sigma^2)]}{\Pr[\theta_i | \mathcal{N}(0, \sigma^2)]} \\
  &= \prod_{i = 1}^T \left( q\frac{\Pr[\theta_i | \mathcal{N}(1, \sigma^2)]}{\Pr[\theta_i | \mathcal{N}(0, \sigma^2)]} + 1 - q \right)
\end{align*}

For numerical stability, we equivalently threshold $\log (\Lambda(\theta)) = \sum_{i = 1}^T \log \left( q\frac{\Pr[\theta_i | \mathcal{N}(1, \sigma^2)]}{\Pr[\theta_i | \mathcal{N}(0, \sigma^2)]} + 1 - q \right)$ instead.
For conciseness, we let $L(\theta_i) = \log \left( q\frac{\Pr[\theta_i | \mathcal{N}(1, \sigma^2)]}{\Pr[\theta_i | \mathcal{N}(0, \sigma^2)]} + 1 - q \right)$ and let the sum be $L_k = \sum_{i = 1}^k L(\theta_i)$.
One key thing to note here is that the likelihood ratios of each individual iterate $L(\theta_i)$ are independent of the other iterates.
This enables us to construct a loss function that performs this likelihood ratio test at each iterate individually and aggregates them over multiple steps.

\subsection{Constructing our loss function}
\label{sec:loss}
Now, we move on to constructing our loss function.
To that end, we first observe that the loss function is only used to compute the gradient $\nabla \tilde{\ell}$, and therefore, we directly construct this gradient function ($\tilde{g} = \nabla \tilde{\ell}$) instead.
Subsequently, our gradient function consists of 3 steps:
\begin{enumerate}
  \item Decode previous iterate to the partial sum of likelihood ratios and previous value, i.e., $\text{Decode}(\theta_k) = (L_{k - 1}, v_k)$.
  \item Perform likelihood ratio test on $v_k$ i.e., $L(v_k)$.
  \item Re-encode the likelihood ratio test and remove the raw value of $v_k$, i.e., $\text{Encode}((L(v_k), -v_k))$.
\end{enumerate}

As we have already shown how to perform the likelihood ratio test in the previous section, what remains is to design appropriate Encode and Decode functions.
There are two main considerations when designing these functions.
Firstly, the encoding should not be corrupted by the addition of noise and other gradients that happen in the update rule.
To do so, we encode the partial sum of likelihood ratios into the higher digits (e.g., 10s or 100s) outside of the range of the other gradients and noise (w.h.p).
Secondly, the encoding cannot be too large, or it will be clipped using the gradient clipping function.
To combat this, we aggregate the encoding over a large number of samples, such that even though each individual gradient is small, when added together, they will reconstruct the original encoding.
Subsequently, the loss function we use is given in Algorithm~\ref{alg:gradient}.
Observe that the loss function we construct $\tilde{\ell}$ is non-convex.

Note that the loss function now depends on the sampling rate $q$ and noise multiplier $\sigma$, which can be assumed to be available to the loss function, as they are global non-sensitive hyper-parameters.
$N$ is the expected size of the dataset to be sampled at each iteration (i.e., $N = q|D|$), which will not ``break'' DP as long as the same value is used for both neighboring datasets $D$ and $D'$ (in practice, we set $N$ to be the expected data size for the smaller of the neighboring datasets).
Lastly, depending on how large $\sigma$ is, the encoding is generic and can be adjusted to encode the likelihood ratio into the 10s, 100s, or 1000s.
In practice, we use the ``68-95-99.7'' rule that states that 99.7\% of samples from the normal distribution with mean $\mu$ and standard deviation $\sigma$ lie within the $\mu \pm 3\sigma$ range.
Therefore, we encode the likelihood ratio sum to the closest power of 10 above $3\sigma$.

\begin{algorithm}[!t]
  \small
  \caption{Our gradient loss function ($\tilde{g} = \nabla \tilde{\ell}$)}\label{alg:gradient}
  \begin{algorithmic}[1]
    \Require Sample, $x$. Previous iterate, $\theta_k$.
    \LineComment Not first iterate
    \If{$\theta_k = 0$}
      \State \Return $x$
    \EndIf
    \LineComment Decode previous iterate
    \State $\underline{L_{k - 1}} \leftarrow \text{round }\theta_k\text{ to nearest 10}$
    \State $v_k \leftarrow \theta_k - \underline{L_{k - 1}}$
    \LineComment Perform likelihood ratio test
    \State $L(v_k) \leftarrow \log \left( q\frac{\Pr[v_k | \mathcal{N}(1, \sigma^2)]}{\Pr[v_k | \mathcal{N}(0, \sigma^2)]} + 1 - q \right)$
    \LineComment Encode 2 d.p. value of likelihood ratio test in the 10s
    \State $\underline{L(v_k)} \leftarrow \lceil L(v_k) * 100 \rfloor * 10$
    \State \Return $(\underline{L(v_k)} - v_k) / N + x$
  \end{algorithmic}
\end{algorithm}

\subsection{Distinguishing the outputs of DP-SGD}
The last question that remains to be answered is ``how does an adversary distinguish between $\theta_T \leftarrow \text{DP-SGD}(D; \tilde{\ell}, \cdot)$ and $\theta_T' \leftarrow \text{DP-SGD}(D'; \tilde{\ell}, \cdot)$?''.
To do so, the adversary runs the gradient loss function one last time on $\theta_T$ and $\theta_T'$ and extracts the (full) likelihood ratio sum, i.e., $o = (\theta_T + N * \tilde{g}(0, \theta_T)) / 1000 \approx L_T$ and $o' = (\theta_T' + N * \tilde{g}(0, \theta_T')) / 1000 \approx L_T'$.
What the adversary is left with is approximately the result of the likelihood ratio test performed on $(v_1,...,v_T)$ and $(v_1',...,v_T')$.
Crucially, note that the adversary only needs access to the final iterate to perform this final extraction step.
Therefore, for our (non-convex) loss function, distinguishing between the final iterates is equivalent to distinguishing between all iterates.

\subsection{Auditing DP-SGD}
\label{sec:audit}
Although our loss function is theoretically designed to make the final iterate of DP-SGD as distinguishable as the sequence of all iterates, in our work, we verify this empirically as well by \textit{auditing} DP-SGD with our loss function.
Here, we briefly explain the method we use to audit and show the detailed algorithm in Algorithm~\ref{alg:audit}.

First, we fix neighboring datasets $D$ and $D'$ and run DP-SGD with our loss function repeatedly on $D$ and $D'$.
Next, the outputs are made more distinguishable by extracting the full likelihood ratio sum, as explained above.
The likelihood ratio sum is then threshold-ed to generate an \textit{observed} FPR-FNR curve.

Subsequently, to derive an empirical estimate $\varepsilon_{emp}$, we first approximate the trade-off function for DP-SGD (with composition) using PLD at regular (0.1) intervals of $\varepsilon$s in the range $[0.5, 20.0]$.
Next, we compare the observed FPR-FNR curve with the predicted trade-off functions from PLD.
Specifically, we output the $\varepsilon_{emp}$ for which the trade-off function predicted by PLD most closely matches (but does not exceed) the observed FPR-FNR curve.

Finally, if we observe that $\varepsilon_{emp} \approx \varepsilon$, then the privacy guarantees of hidden state DP-SGD at $\varepsilon$ is equivalent to the privacy guarantees of DP-SGD with composition at $\varepsilon_{emp}$.
Therefore, we can conclude that there can be no hidden state privacy amplification for DP-SGD for general loss functions.

\begin{algorithm}[!t]
  \small
  \caption{Auditing DP-SGD with our loss function $\tilde{\ell}$}\label{alg:audit}
  \begin{algorithmic}[1]
  \Require Neighboring inputs, $D, D'$. Loss function, $\tilde{\ell}$. Initial model parameters, $\theta_0$. Learning rate, $\eta$. Gradient clipping norm, $C$. Noise multiplier, $\sigma$. Sampling rate, $q$. Number of steps, $T$. Number of repetitions, $2R$.
  \LineComment Generate observations from the final iterate of DP-SGD
  \State Observations $O \leftarrow \{\}$, $O' \leftarrow \{\}$
  \For{$r \in [R]$}
      \State $N \leftarrow q|D|$
      \State $\theta_T \leftarrow \text{DP-SGD}(D; \tilde{\ell}, \theta_0, \eta, C, \sigma, q, T)$
      \State $\theta'_T \leftarrow \text{DP-SGD}(D'; \tilde{\ell}, \theta_0, \eta, C, \sigma, q, T)$
      \State $O[t] \leftarrow \theta_T + (N * \tilde{g}(0, \theta_T))/1000$
      \State $O'[t] \leftarrow \theta'_T + (N * \tilde{g}(0, \theta_T))/1000$
  \EndFor

  \\
  \LineComment Calculate the observed FPR-FNR curve
  \State $\text{FNRs} \leftarrow \{\}$
  \For{$\tau \in O \cup O'$}
    \State $\alpha \leftarrow |\{o | o \in O, o \geq \tau \}| / |O|$
    \State $\beta \leftarrow |\{o | o \in O', o < \tau \}| / |O'|$
    \State $\text{FNRs}[\alpha] \leftarrow \beta$
  \EndFor

  \\
  \LineComment Estimate empirical $\varepsilon_{emp}$
  \State $\Upsilon \leftarrow \{0.5, 0.6, ..., 20.0\}$
  \For{$\hat{\varepsilon} \in \Upsilon$}
    \For{$\alpha, \beta \in \text{FNRs}$}
      \LineComment Approximate trade-off from PLD
      \State $\hat{\beta} \leftarrow \text{PLD}(\hat{\varepsilon})(\alpha)$
      \LineComment Observed trade-off violates predicted trade-off function
      \If{$\beta < \hat{\beta}$}
        \State Skip to next $\hat{\varepsilon}$
      \EndIf
    \EndFor
    \State \Return $\varepsilon_{emp} \leftarrow \hat{\varepsilon}$
  \EndFor
  \end{algorithmic}
\end{algorithm}

\section{Experiments}
\begin{figure*}[t]
  \centering
  \includegraphics[width=\linewidth]{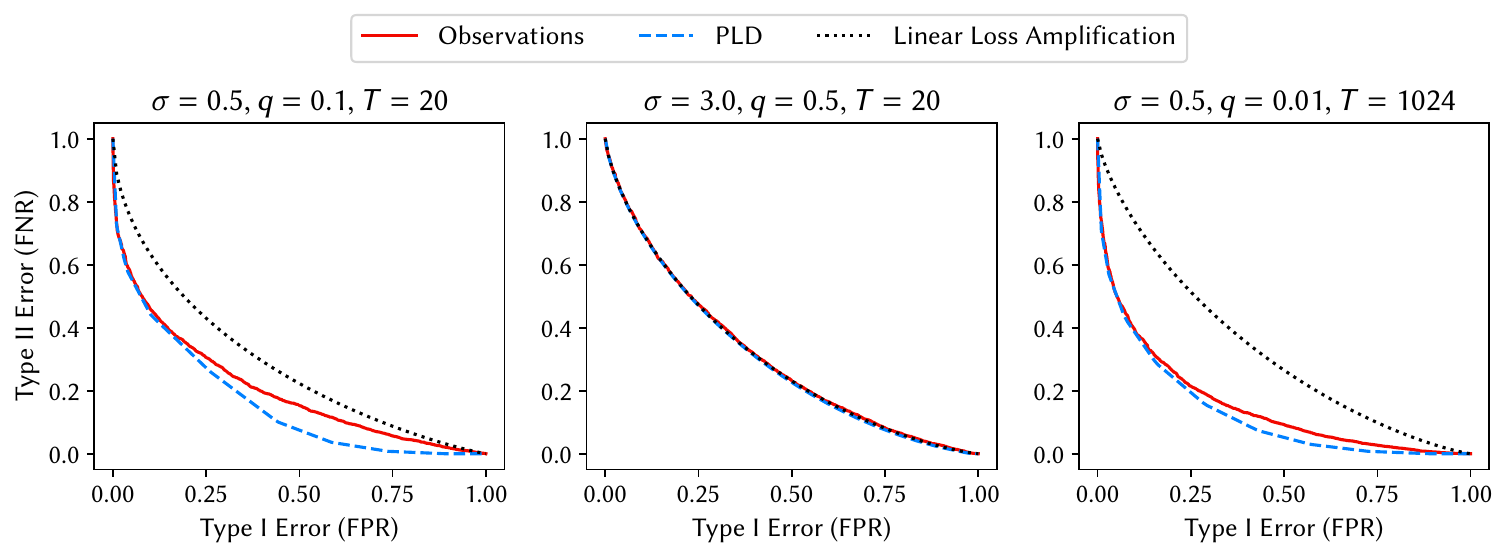}
  \caption{Comparing FPR-FNR curve observed by thresholding final iterate of DP-SGD with our loss function (Observations) with predicted trade-off function from PLD when all iterates are released for DP-SGD (PLD) and trade-off function predicted by PLD for hidden state DP-SGD with linear loss~\cite{choquette-choo2024privacy} (Linear Loss Amplification).}
  \label{fig:compare_tradeoffs}
\end{figure*}

\begin{figure*}[t]
  \centering
  \includegraphics[width=\linewidth]{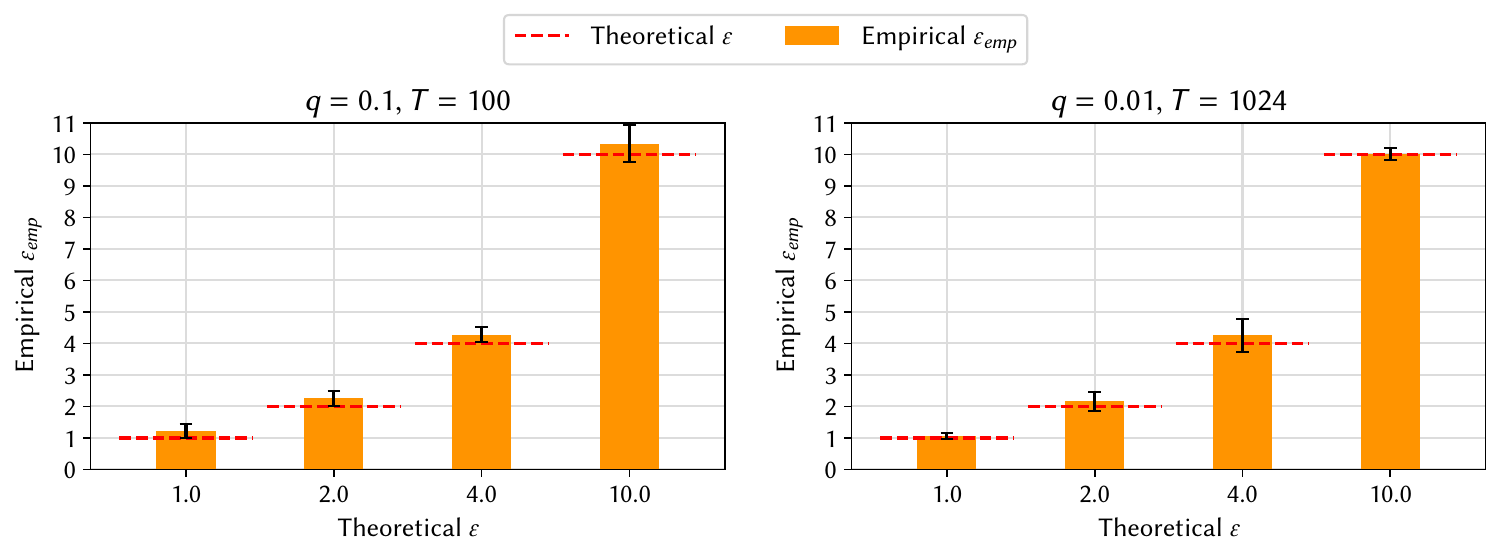}
  \caption{Auditing the final iterate of DP-SGD with our loss function for varying $\varepsilon = 1.0, 2.0, 4.0, 10.0$ and in different settings ($q = 0.1, T = 100$ and $q = 0.01, T = 1024$). Error bars are $\pm 2\sigma$.}
  \label{fig:audit}
\end{figure*}

In this section, we empirically verify that for our loss function (defined in Section~\ref{sec:loss}), distinguishing the final iterate of DP-SGD (hidden state) is equivalent to distinguishing all iterates.
To that end, we first construct worst-case neighboring datasets $D = \{0,...,0\}$ s.t. $|D| = $10B and $D' = D \cup \{1\}$.
Although the neighboring datasets we use seem degenerate, we note that this is acceptable since the privacy guarantees of DP-SGD (or any DP algorithm in general) should hold for all neighboring datasets, including worst-case ones.
In this case, a dataset with a large number of `0's is necessary to reliably encode the likelihood ratio sum.
However, we note that it is trivial to extend our loss function to support any sufficiently large dataset, e.g., LAION-5B.
Then we run DP-SGD with our loss function on $D$ and $D'$ 10k times in total (5k for each dataset), which we use to report FPR-FNR curves and derive empirical $\varepsilon_{emp}$ values.
Additionally, to derive the empirical $\varepsilon_{emp}$, we average the empirical estimate achieved over five independent runs.
All experiments were run on a single server with an Intel Core i7 CPU with 12 cores and 32GB of RAM.

\subsection{Comparing FPR-FNR curves}
We first begin by comparing the observed FPR-FNR curves from distinguishing the last iterate of DP-SGD (with our loss function) with the trade-off curve predicted by PLD, which corresponds to releasing all iterates of DP-SGD.
To provide further context, we additionally plot the trade-off function for DP-SGD with linear loss, which is expected to have hidden state privacy amplification~\cite{choquette-choo2024privacy}.
More precisely, we plot the approximate trade-off function for the Mixture of Gaussians mechanism, which has equivalent privacy guarantees achieved by releasing only the final iterate of DP-SGD initialized with a linear loss function.

Subsequently, in Figure~\ref{fig:compare_tradeoffs}, we plot the corresponding trade-off functions for three different hyper-parameters covering the range of noise multipliers ($\sigma$), sampling rates ($q$), and steps ($T$).
First, we notice that regardless of the configuration of hyper-parameters used, the FPR-FNR curve observed for the final iterate of DP-SGD with our loss function almost exactly matches the predicted trade-off function of PLD.
Although in some cases, the observed FNR at large FPRs appears to be larger than the predicted FNR from PLD, we note that this is because the trade-off function approximated from PLD is \textit{symmetric} as explained in Section~\ref{sec:tradeoff}.
In fact, if the neighboring datasets used are swapped, i.e., $D' = \{0,...,0\}$ s.t. $|D'| =$10B and $D = D' \cup \{1\}$, the observed FPR-FNR curve will be the inverse of what we see in Figure~\ref{fig:compare_tradeoffs}, which will correspond to the FNRs predicted by PLD at high FPRs.

Second, we observe that even when there is a large hidden state privacy amplification expected, e.g., $\sigma = 0.5, q = 0.01, T = 1024$, the observed FPR-FNR curve for the final iterate of DP-SGD with our loss function deviates from this amplification significantly.
This further reinforces the fact that DP-SGD, with our loss, does not experience any hidden state amplification even though only the final iterate has been released.

\subsection{Auditing results}
On top of comparing the trade-off functions visually, we also rigorously audit the final iterate of DP-SGD with our loss function using the method explained in Section~\ref{sec:audit}.
To that end, in Figure~\ref{fig:audit}, we plot the empirical $\varepsilon_{emp}$s obtained for varying theoretical $\varepsilon$s for two sets of hyper-parameters.
We can see clearly that the empirical $\varepsilon_{emp}$ matches the theoretical $\varepsilon$ exactly for all settings.
We note that although the empirical privacy estimate appears to slightly exceed the theoretical guarantee, this is expected since we do not compute confidence intervals for the observed FPR-FNR curve, and in fact, the true theoretical $\varepsilon$ falls within $\pm 2\sigma$ of the empirical guarantees achieved.
Therefore, we observe that the current privacy analysis of DP-SGD is indeed tight with respect to general loss functions, even when only the final iterate is released.

\section{Conclusion}
In this work, we studied whether there can be a hidden state privacy amplification result for DP-SGD for all non-convex loss functions in general.
To that end, we first constructed a worst-case loss function for DP-SGD, which stores the information of all iterates into the final iterate.
Next, we verified that the empirical privacy leakage from the final iterate of DP-SGD initialized with our worst-case loss function matches the theoretical privacy analysis of DP-SGD, which assumes that all iterates are released.
This shows that the privacy guarantees of DP-SGD with our worst-case loss function cannot be amplified on the basis that only the final iterate is released.
Hence, our loss function acts as a counter-example, and we conclude that no hidden state privacy amplification results are possible for DP-SGD for all non-convex loss functions in general.

\paragraph{Future Work}
Our main result is that hidden state privacy amplification results for DP-SGD are not possible for all non-convex loss functions (incl. worst-case) in general.
However, it is currently unclear if our results apply specifically to more commonly used (\textit{natural}) non-convex loss functions in differentially private machine learning, e.g., ReLU activations with Cross Entropy loss.
For instance, it is possible that such \textit{natural} non-convex loss functions satisfy additional properties that are currently unknown, which similarly result in hidden state privacy amplification.
Therefore, one remaining open challenge will be to investigate whether it is possible to extract the same level of information from the final iterate of DP-SGD when used together with \textit{natural} loss functions as well.

\begin{acks}
  This work has partly been supported by a National Science Scholarship (PhD) from the Agency for Science Technology and Research, Singapore (A*STAR).
  We wish to thank Emiliano De Cristofaro and Jamie Hayes for providing ideas and feedback throughout the project.
  We also wish to thank Georgi Ganev for their feedback on early versions of the paper.
\end{acks}

\printbibliography

\appendix

\end{document}